\documentclass[10pt,journal,twoside,web]{ieeecolor}
\usepackage{generic}
\usepackage[T1]{fontenc}
\usepackage[utf8]{inputenc}
\usepackage{cite}
\usepackage{amsmath,amssymb,amsfonts,mathtools}
\usepackage{graphicx}
\usepackage{booktabs,array,multirow,tabularx}
\usepackage{placeins}
\usepackage{float}
\usepackage{dblfloatfix}
\usepackage{textcomp}
\usepackage{url}
\usepackage{microtype}
\usepackage{xcolor}
\definecolor{wordblue}{RGB}{0,0,0}
\definecolor{wordred}{RGB}{255,0,0}
\definecolor{nblue}{RGB}{0,0,0}
\definecolor{subsectioncolor}{RGB}{0,0,0}
\providecommand{\refname}{References}
\makeatletter
\let\NAT@parse\undefined
\makeatother
\usepackage{hyperref}
\hypersetup{
  colorlinks=true,
  linkcolor=wordblue,
  citecolor=wordblue,
  urlcolor=black,
  pdfborder={0 0 0}
}
\providecommand{\significancenote}[1]{#1}
\newcolumntype{Y}{>{\centering\arraybackslash}X}
\newcolumntype{L}{>{\raggedright\arraybackslash}X}
\makeatletter
\newcommand{\currenttablelabel}[1]{%
  \phantomsection
  \edef\@currentlabel{\thetable}%
  \label{#1}}
\def\fnum@figure{\color{wordblue}\sf Fig.~\thefigure}
\makeatother
\AddToHookNext{shipout/after}{\global\hfuzz=0.1pt}

\title{Mixed-Sample Data Augmentation Based on a Low-Dimensional Semantic Manifold for Tumor Classification in Medical Images}
\author{Fangqing Liu, \IEEEmembership{member, IEEE}, Ruqing Lv, Guosheng Liu, Xueming Yan \IEEEmembership{Senior Member, IEEE}, and Han Huang*, \IEEEmembership{Senior Member, IEEE}
\thanks{This work is supported by National Natural Science Foundation of China (62576116, 92570116).
F. Liu is with the School of Management, Guangdong University of Technology, Guangzhou, 510520 China, Email: peerfog@gdut.edu.cn.
R. lv is with the School of Future Technology, South China University of Technology, Guangzhou, 510641 China, Email: lrq20060209@qq.com.
G. Liu is with the School of Management, Guangdong University of Technology, Guangzhou, 510520 China, Email: gliu@gdut.edu.cn.
X. Yan is with the School of Information Science and Technology, Guangdong University of Foreign Studies, Guangzhou, 510006 China, Email: yanxm@gdufs.edu.cn.
H. Huang is with the School of Software Engineering, Sun Yat-sen University, Zhuhai, 519082 China. Email: huangh985@mail.sysu.edu.cn}}
\begin{document}
\maketitle

\begin{abstract}
Tumor classification in medical images is limited by costly annotation and scarce high-quality data. Local mixing effectively transfers discriminative regions without dense annotations. However, under image-level supervision, source-region selection is sensitive to response variations, and unconstrained placement can discard target evidence. To address these issues, we propose a mixed-sample data augmentation method based on a low-dimensional semantic manifold (MSDA-LDSM). To extract repeatable source evidence, a stable semantic anchor is constructed by integrating class-response intensity and variability under dropout-based stochastic subnetworks. To preserve target semantics, the source region is fixed, which reduces the placement search to a constrained two-dimensional space on the manifold. Candidate placements are filtered by target-response retention and ranked using semantic, structural, and spatial scores within a fixed budget. The mixed label is computed from the retained class-response fractions of both images. Experiments on three datasets demonstrate an average accuracy gain of 7.81 percentage points over the baseline and a mean accuracy rank of 1.33 among state-of-the-art methods. Ablation studies confirm that variability-aware anchor construction effectively extracts repeatable local evidence, while the constrained placement procedure preserves target semantics under a limited computational budget, thereby limiting redundant exploration.
\end{abstract}


\section{Introduction}
\label{sec:introduction}

Malignant tumors continue to pose a major global health burden, with 20.6 million new cases and 9.8 million deaths in 2024 \cite{ref01}. Computed tomography (CT), magnetic resonance imaging (MRI), and histopathology provide complementary diagnostic information. Deep neural networks extract discriminative features from these data \cite{ref02}. However, costly acquisition and annotation limit the scale of tumor datasets, making data augmentation essential for alleviating overfitting and enhancing generalization \cite{ref04}.

Existing augmentation methods can be grouped into four categories: predefined transformations, sample-based recombination, learning-driven search, and semantic-guided generation. Predefined transformations are widely used for basic augmentation, whereas mixed-sample methods, such as MixUp \cite{ref05}, CutMix \cite{ref06}, and SnapMix \cite{ref07}, generate new images by recombining existing samples. Learning-driven methods such as AutoAugment \cite{ref08} automate policy search and reduce manual design effort, whereas semantic-guided methods such as TextManiA \cite{ref09} incorporate external semantics to guide augmentation. Within these families, medical-image-specific variants further incorporate anatomical constraints \cite{ref10}, and tumor-oriented variants account for domain shift and lesion variability \cite{ref11,ref12}. Among these, local mixing is particularly relevant to image-level tumor classification because it transfers selected discriminative regions without requiring dense annotations.

Local mixing requires two coupled decisions: source-region selection and target placement. First, a single response may be unreliable or sensitive to perturbations \cite{ref13}, hindering the acquisition of repeatable source evidence. Second, placement must account for lesion location and shape while preserving target semantics and local structure \cite{ref14}. These challenges motivate two questions: how to extract repeatable local evidence under image-level supervision, and how to constrain target placement within a limited computational budget. Our first hypothesis, \textbf{H1}, is that responses with high mean and low variability under stochastic subnetwork perturbations provide repeatable class evidence. Our second hypothesis, \textbf{H2}, is that this evidence can serve as a semantic anchor for inserted content. With the source region fixed, candidate placements are represented by two translation coordinates. Placements are filtered by target-response retention and ranked using semantic, structural, and distance scores within a fixed candidate budget.

Based on these hypotheses, we propose a mixed-sample data augmentation method based on a low-dimensional semantic manifold (MSDA-LDSM). For H1, a stable-response map is constructed from class-response intensity and variability, from which the source-region mask is extracted. For H2, the source region is held fixed, and placement is parameterized by horizontal and vertical translation coordinates. Candidate placements are evaluated using target-response retention and structural priors. Response-based labels connect the retained content to supervision. Experiments on ChestCT \cite{ref15}, BrainTumor \cite{ref16}, and BreakHis \cite{ref17} with ResNet-18/50 \cite{ref18} and GoogLeNet \cite{szegedy2015googlenet} assess anchor repeatability, placement constraints, and classification generalization.

The main contributions of this work are summarized as follows.
\begin{enumerate}
  \item We develop a procedure for constructing stable semantic anchors from class-response intensity and variability under image-level supervision. By integrating response intensity with variability under stochastic subnetwork perturbations, this procedure enables the identification of repeatable local class evidence and yields an explicit criterion for anchor construction, as stated in H1.
  \item We formulate anchor-guided augmentation as a constrained placement problem in a two-dimensional search space on a low-dimensional semantic manifold, as stated in H2. The placement is jointly determined by a fixed source region, target-response retention, and structural and spatial priors. A fixed candidate budget restricts the number of evaluated placements, thereby limiting redundant exploration. Response-based labels are also derived to connect these spatial decisions with image-level supervision.
\end{enumerate}

The rest of this paper is organized as follows. Section~\ref{sec:background} presents the optimization formulation and related work. Section~\ref{sec:method} details MSDA-LDSM, Section~\ref{sec:experiments} reports the experiments, and Section~\ref{sec:conclusion} concludes the paper.

\section{Background}
\label{sec:background}

This section formulates data augmentation for image-level tumor classification as a mixed-variable black-box optimization problem. It then reviews four categories of related methods and identifies the research gap addressed by MSDA-LDSM.

\subsection{Optimization Formulation of Data Augmentation for Tumor Classification in Medical Images}
\label{sec:optimization}

This study considers image-level tumor classification without pixel-level lesion annotations. Let \(\mathbf{I}\) and \(y\) denote an image and its image-level label, and let \(N_{\mathrm{tr}}\) and \(N_{\mathrm{val}}\) be the numbers of training and validation samples. The corresponding datasets are

\begin{equation}
\footnotesize
\begingroup\thinmuskip=1mu\medmuskip=1mu\thickmuskip=2mu
\mathcal{D}_{\mathrm{tr}}=\{(\mathbf{I}_i,y_i)\}_{i=1}^{N_{\mathrm{tr}}},\quad
\mathcal{D}_{\mathrm{val}}=\{(\mathbf{I}^{\mathrm{val}}_j,y^{\mathrm{val}}_j)\}_{j=1}^{N_{\mathrm{val}}}.
\label{eq:datasets}
\endgroup
\end{equation}

An augmentation policy is denoted by \(\boldsymbol{\alpha}=(\alpha_d,\alpha_c)\). The discrete component \(\alpha_d\) encodes discrete decisions, such as operation type and sample pairing. The continuous component \(\alpha_c\) contains continuous parameters, such as magnitude and mixing weight. Let \(A_{\boldsymbol{\alpha}}\) denote the augmentation mapping. For each ordered sample group \(s_k\), define \((\widetilde{\mathbf{I}}_k,\widetilde{\mathbf{y}}_k)=A_{\boldsymbol{\alpha}}(s_k)\). The augmented dataset consists of \(N_{\boldsymbol{\alpha}}\) such outputs:

\begin{equation}
\footnotesize
\mathcal{D}_{\mathrm{aug}}(\boldsymbol{\alpha})=
\{(\widetilde{\mathbf{I}}_k,\widetilde{\mathbf{y}}_k)\}_{k=1}^{N_{\boldsymbol{\alpha}}},
\label{eq:augdata}
\end{equation}
where \(N_{\boldsymbol{\alpha}}\) is the number of augmented outputs. Let \(f_{\boldsymbol{\omega}^{*}(\boldsymbol{\alpha})}\) denote the classifier trained under a fixed protocol using \(\mathcal{D}_{\mathrm{aug}}(\boldsymbol{\alpha})\). Its validation performance is

\begin{equation}
\footnotesize
\begingroup\thinmuskip=1mu\medmuskip=1mu\thickmuskip=2mu
P(\boldsymbol{\alpha})=\frac{1}{N_{\mathrm{val}}}\sum_{j=1}^{N_{\mathrm{val}}}
\mathbb{I}\!\left[\underset{c\in\{1,\ldots,C\}}{\arg\max}\,
f_{\boldsymbol{\omega}^{*}(\boldsymbol{\alpha})}(\mathbf{I}^{\mathrm{val}}_j)_c=y^{\mathrm{val}}_j\right].
\label{eq:performance}
\endgroup
\end{equation}

In \eqref{eq:performance}, \(C\) is the number of classes, and \(f_{\boldsymbol{\omega}^{*}(\boldsymbol{\alpha})}(\mathbf{I}^{\mathrm{val}}_j)_c\) is the classifier output for class \(c\). The notation \(\mathbb{I}[\cdot]\) denotes the indicator function, and \(y_j^{\mathrm{val}}\) is the validation class index. Since each evaluation of \(P(\boldsymbol{\alpha})\) requires costly classifier training and validation with no directly usable analytical form, the augmentation problem is formulated as a mixed-variable black-box optimization:

\begin{equation}
\footnotesize
\boldsymbol{\alpha}^{*}\in
\underset{\boldsymbol{\alpha}\in\mathcal{A}_{\mathrm{feas}}}{\arg\max}\
P(\boldsymbol{\alpha}).
\label{eq:policy}
\end{equation}
Here, \(\mathcal{A}_{\mathrm{feas}}\) denotes the set of feasible policies that satisfy task-specific validity constraints and remain within the computational budget. 

The formulation in \eqref{eq:augdata} accommodates single-sample transformations and cross-sample augmentation. The resulting labels may be one-hot or soft. For online augmentation, \(A_{\boldsymbol{\alpha}}\) is invoked during mini-batch optimization. The validation set is used for policy evaluation, and the test set is reserved for final evaluation. 

\subsection{Related Work}
\label{sec:related}

This section reviews four general categories of data augmentation methods for tumor classification in medical images: predefined transformations, sample-based recombination, learning-driven search, and semantic-guided generation. It then discusses medical-image-specific and tumor-oriented variants as application-specific extensions of these categories.

Predefined transformations are widely used for basic augmentation. Sample-based recombination, particularly mixed-sample augmentation, has progressed from global interpolation to localized, semantic-guided recombination. MixUp \cite{ref05} and Manifold Mixup \cite{ref19} interpolate inputs or features, whereas CutMix \cite{ref06} and FMix \cite{ref20} replace local regions. Later methods improve region selection, alignment, and label estimation via saliency or feature correspondence \cite{ref07,ref21,ref22,ref23,ref24,ref25}, while Simple Copy-Paste \cite{ref26} extends recombination to annotated instances. These methods demonstrate the value of local mixing. However, they do not directly optimize placement under medical structural constraints. They select mixing regions using deterministic evidence or instance masks, without directly evaluating biological plausibility during placement.

Learning-driven search reduces manual design effort. AutoAugment \cite{ref08}, RandAugment \cite{ref27}, and SRA \cite{ref28} represent the progression from policy search to reduced or sample-aware control. Semantic-guided generation uses sample or language information to guide augmentation. DiffuseMix \cite{ref29}, TextManiA \cite{ref09}, and ALIA \cite{ref30} use generative editing, textual attributes, or semantic filtering. These methods primarily optimize global policies, feature perturbations, or generated content. They rarely achieve repeatable, localized placement of tumor lesions, a property essential for augmentation under medical structural constraints.

Beyond these four general categories, medical-image-specific variants incorporate anatomical, domain-variation, and class-imbalance constraints. MedAugment \cite{ref10} restricts transformation types and magnitudes. AADG \cite{ref31}, JCSAugment \cite{ref32}, and Balanced-MixUp \cite{ref33} adapt policies to domain generalization, segmentation, or class imbalance. BCP \cite{ref34} introduces bidirectional regional recombination for semi-supervised segmentation. StyCona \cite{ref35} varies domain style while preserving semantic content. These methods encode useful medical priors, yet most of them target segmentation, domain adaptation, or full-image transformations. Image-level tumor mixing remains underexplored.

Tumor-oriented variants operate at multiple granularities, including histopathology staining, bag-level augmentation, representation-level augmentation, and lesion-level augmentation. For histopathology, STRAP \cite{ref11}, HistAuGAN \cite{ref36}, and RandStainNA \cite{ref37} reduce staining variation across centers. RankMix \cite{ref38}, HistAug \cite{ref39}, multimodal online augmentation \cite{ref40}, SSRDL \cite{ref41}, and DPBAug \cite{ref42} operate on bags, representations, or online regional content. Among these, lesion-level methods transfer or synthesize localized tumor regions. TumorCP \cite{ref43}, CarveMix \cite{ref14}, InsMix \cite{ref44}, and SelfMix \cite{ref45} use masks or instance regions. RBTCA \cite{ref46} further incorporates textual constraints. DiffTumor \cite{ref12} and text-conditioned nuclei synthesis \cite{ref47} adopt diffusion models to generate tumor regions.

To summarize, existing augmentation methods increase diversity through predefined transformations, sample-based recombination, learned policies, or semantic-guided generation, together with application-specific medical and tumor constraints. Coordinating repeatable region selection and constrained placement remains challenging under image-level supervision. This motivates MSDA-LDSM, a closed-loop framework for microscale searching that bridges repeatable local class evidence and constrained placement within a low-dimensional semantic manifold.

\section{Method}
\label{sec:method}

MSDA-LDSM addresses the two hypotheses introduced in Section~\ref{sec:introduction} using a common response representation. For H1, the method identifies local class evidence characterized by high mean response and low variability across stochastic subnetworks. For H2, the source region is fixed by the extracted mask. Candidate placements are filtered by target-response retention and ranked using semantic, structural, and distance scores. Response-based label weights are computed to connect these spatial decisions with image-level supervision. The following subsections explain how each component addresses H1 and H2.

\subsection{Overall Framework of MSDA-LDSM}
\label{sec:framework}

\begin{figure*}[!tb]
\centering
\includegraphics[width=0.780\textwidth]{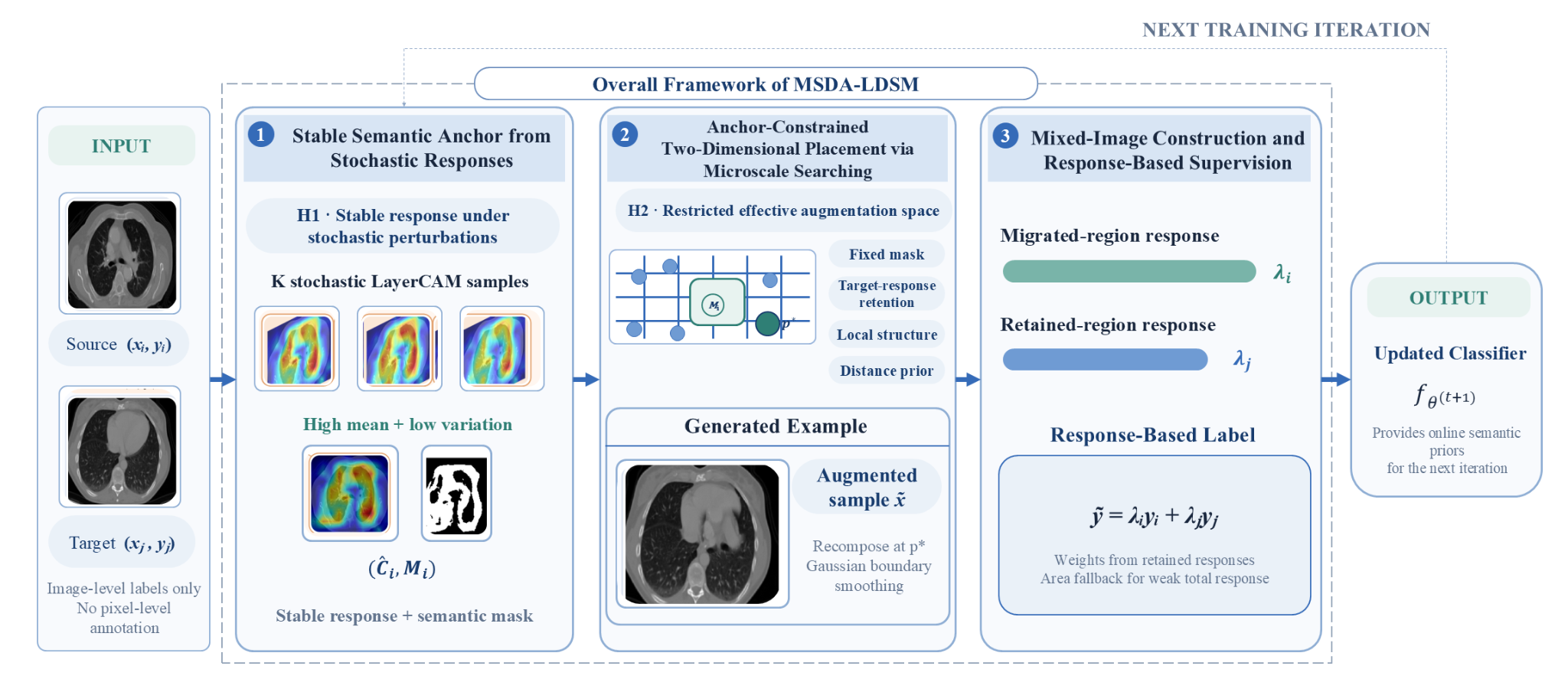}
\caption{Overall framework of MSDA-LDSM. Under H1, a stable semantic map and a source-region mask are constructed from response statistics. Under H2, placement is parameterized by two translation coordinates with the source mask held fixed. Candidate selection is guided by target-response retention, as well as structural and spatial priors. Response-based labels and regional feature consistency connect sample construction to classifier training.}
\label{fig:framework}
\end{figure*}

Figure~\ref{fig:framework} shows MSDA-LDSM in the context of online classifier training. Let $(x_i,y_i)$ and $(x_j,y_j)$ denote the source and target image-label pairs, respectively. The source image $x_i$ provides the region to be transferred, and the target image $x_j$ receives it. The labels $y_i$ and $y_j$ are encoded as one-hot vectors. For each image, a normalized stable-response map represents class evidence, and a binary mask identifies the semantic region. The source mask determines the source content to be transferred. A target-response retention criterion is applied to exclude candidate placements that would remove excessive target class evidence. The same response representation is also used to compute the label weights, thereby connecting region selection, placement, and supervision.

The manifold hypothesis suggests that meaningful variations in high-dimensional data concentrate on lower-dimensional structures \cite{ref48}. To address H1, a normalized stable-response map is constructed using response statistics from stochastic subnetworks, and a binary source-region mask is extracted from this map. With the source mask fixed, only the horizontal and vertical translation coordinates remain variable. This anchor-defined placement space is referred to as the low-dimensional semantic manifold under H2.

Microscale searching is applied to evaluate a fixed number of candidate placements in this two-dimensional decision space \cite{ref49}. A target-response retention constraint is used to exclude placements that would remove excessive target evidence. The remaining candidates are ranked using semantic, structural, and distance scores. This yields explicit placement criteria under a fixed candidate budget.

\subsection{Stable Semantic Anchor from Stochastic Responses}
\label{sec:prior}

H1 states that a local class response with high mean intensity and low variability across stochastic subnetworks provides repeatable evidence for the image-level class. A single response map cannot capture both properties. Therefore, an empirical response distribution is estimated at each spatial location, using its mean to represent response intensity and its standard deviation to represent sensitivity to perturbations.

\textbf{1) Stochastic Response Sampling:} To characterize the variability of class evidence, we sample multiple response maps under stochastic subnetwork perturbations. We use LayerCAM \cite{ref50} to produce a spatial attribution map for the class indicated by the image-level label. Dropout \cite{ref51} is enabled at a high-level convolutional stage, such that each forward pass corresponds to a different subnetwork. The input image, target class, and network parameters remain fixed while the dropout mask changes. The resulting $K$ maps, denoted by $S_i^{(k)}$, capture response variability induced by dropout. A deterministic map $S_i^{\mathrm{det}}$ is obtained with dropout disabled. Both types of response are extracted from the same network layer.

\textbf{2) Response Statistics and Semantic Anchor Construction:} The repeated response samples are summarized into empirical statistics at each spatial location $(u,v)$:

\begin{equation}
\footnotesize
\mu_i(u,v)=\frac{1}{K}\sum_{k=1}^{K}S_i^{(k)}(u,v)
\label{eq:mean}
\end{equation}

\begin{equation}
\footnotesize
\sigma_i(u,v)=\sqrt{\frac{1}{K}\sum_{k=1}^{K}
\left(S_i^{(k)}(u,v)-\mu_i(u,v)\right)^2}
\label{eq:std}
\end{equation}

The mean measures the strength of recurring class evidence, while the standard deviation measures its variation across subnetworks. To incorporate the unperturbed reference, the deterministic response is fused with the stochastic mean:

\begin{equation}
\footnotesize
S_i^{\mathrm{fus}}(u,v)=\eta S_i^{\mathrm{det}}(u,v)+(1-\eta)\mu_i(u,v)
\label{eq:fusion}
\end{equation}

The coefficient \(\eta\) controls the relative weight of the deterministic response. The stability score combines the fused response intensity with stochastic variability:

\begin{equation}
\footnotesize
C_i(u,v)=\frac{S_i^{\mathrm{fus}}(u,v)}{\sigma_i(u,v)+\varepsilon}
\label{eq:stability}
\end{equation}
Here, \(\varepsilon>0\) prevents division by zero. For a fixed response intensity, lower variability increases the stability score. For a fixed variability, a stronger response increases the score. This construction makes the two criteria in H1 explicit in the spatial weights.

\begin{figure*}[!tb]
\centering
\includegraphics[width=0.970\textwidth]{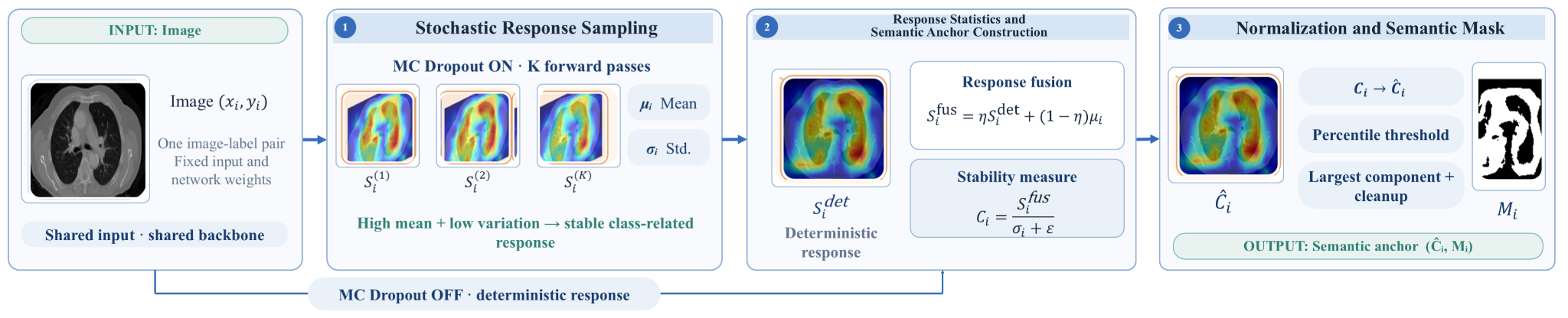}
\caption{Stable semantic anchor from stochastic responses under H1. The deterministic LayerCAM response is fused with the stochastic response mean. The fused response is divided by the empirical response variability to form the stability map. This map is normalized to obtain the continuous stable-response map \(\widehat{C}_i\). Percentile thresholding and component cleanup yield the binary mask \(M_i\).}
\label{fig:prior}
\end{figure*}

The stable-response map is normalized to \(\widehat{C}_i\). The binary mask \(M_i\) is produced by applying a percentile threshold within each image, followed by component cleanup. The percentile rule enables adaptive region selection based on the response values in each image. The mask supplies a fixed region for transfer, while the continuous map retains spatial weights for placement constraints and label construction. Figure~\ref{fig:prior} shows these two complementary outputs. The same procedure is applied to obtain the target map \(\widehat{C}_j\).

\subsection{Anchor-Constrained Two-Dimensional Placement via Microscale Searching}
\label{sec:placement}

H2 concerns how a stable source region can support useful mixing without excessive loss of target evidence. The semantic anchor is instantiated as the fixed source mask \(M_i\), which eliminates shape and scale from the placement decision and leaves only two translation coordinates. These coordinates define the low-dimensional semantic manifold introduced in Section~\ref{sec:framework}. To address H2, the placement decision is decomposed into three components: a fixed-budget candidate domain, a hard retention constraint, and soft preferences over feasible candidates. The retention constraint limits the loss of target response, and the preferences are used to rank feasible placements according to structural and spatial criteria.

\textbf{1) Candidate Domain and Search Budget:} Let \(\Omega\) denote the target image grid and \(M_i^{p}\) the source mask translated to placement coordinate \(p\). The valid domain contains coordinates at which the complete mask lies inside \(\Omega\). The reference coordinate \(p_0\) corresponds to the original location of the source region, with clipping to the valid domain when necessary. With a candidate budget of \(T_{\max}=20\), one candidate is placed at \(p_0\), and the remaining 19 candidates are sampled uniformly:

\begin{equation}
\footnotesize
\begin{aligned}
\Omega_{\mathrm{valid}}&=\{p:\operatorname{supp}(M_i^p)\subseteq\Omega\}\\
\mathcal{P}_{\mathrm{cand}}&=\{p_0\}\cup\{p_t\}_{t=1}^{19},\quad
p_t\sim U(\Omega_{\mathrm{valid}})
\end{aligned}
\label{eq:candidates}
\end{equation}

With \(M_i\) fixed, placement is parameterized by two translation coordinates. The candidate budget limits evaluation to one reference placement and 19 coordinates sampled uniformly from \(\Omega_{\mathrm{valid}}\). Larger distance scores are assigned to placements nearer \(p_0\). The resulting bounded candidate set is used for microscale searching \cite{ref49,ref52}.

\textbf{2) Target Response Retention:} The target-preservation requirement of H2 is expressed as a retention constraint. Let \(z\) index spatial locations, and let \(H_j(z)\) indicate whether \(z\) belongs to the high-response region of the target map \(\widehat{C}_j\). This region is obtained by thresholding as in Section~\ref{sec:prior}. The retained fraction of the target response energy is

\begin{equation}
\footnotesize
\mathcal{R}_j(p)=
\frac{\sum_z H_j(z)\widehat{C}_j(z)[1-M_i^p(z)]}
{\sum_z H_j(z)\widehat{C}_j(z)+\varepsilon}
\label{eq:retention}
\end{equation}
and the candidate set is restricted to
\begin{equation}
\footnotesize
\mathcal{P}_{\mathrm{feas}}=\{p\in\mathcal{P}_{\mathrm{cand}}:\mathcal{R}_j(p)\geq\tau\}
\label{eq:feasible}
\end{equation}
where the threshold \(\tau\) specifies the minimum retained fraction. A candidate placement \(p\) is rejected if \(\mathcal{R}_j(p)<\tau\), regardless of its structural or distance score. The constraint limits the removal of high-response target evidence. It permits partial overlap when the required fraction is retained.

\textbf{3) Structural and Spatial Preferences:} The retention threshold is used to identify feasible placements, but not to determine which of them should be preferred. We therefore define three complementary scores to rank the feasible candidates according to semantic preservation, local structure, and spatial proximity. Let \(\overline{C}_j(p)\) denote the mean normalized target response under \(M_i^p\). The semantic score is designed to favor placements with less target-response energy in the covered region:

\begin{equation}
\footnotesize
\psi_{\mathrm{sem}}(p)=1-\overline{C}_j(p)
\label{eq:semantic-score}
\end{equation}

The structural score is computed from the normalized edge density \(\rho_{\mathrm{ed}}(p)\) and grayscale variance \(\sigma_{\mathrm{gr}}^2(p)\) in the candidate region:
\begin{equation}
\footnotesize
\psi_{\mathrm{str}}(p)=w_e\rho_{\mathrm{ed}}(p)+w_v\sigma_{\mathrm{gr}}^2(p)
\label{eq:structure-score}
\end{equation}
Both descriptors are normalized to a common scale before combination. The weights \(w_e\) and \(w_v\) control the contribution of these two structural descriptors. This score expresses a preference based on local structural complexity. It does not impose a hard condition on anatomical continuity. A spatial prior is applied to favor smaller displacement from \(p_0\):

\begin{equation}
\footnotesize
\psi_{\mathrm{dist}}(p)=\exp\!\left(-\frac{\lVert p-p_0\rVert_2^2}{2\sigma_{\mathrm{dist}}^2}\right)
\label{eq:distance-score}
\end{equation}
Here, \(\sigma_{\mathrm{dist}}\) controls how rapidly the distance preference decays. The structural and distance terms complement the semantic criterion by distinguishing between target contexts with similar response retention. The combined placement objective is

\begin{equation}
\footnotesize
\Phi(p)=\gamma_1\psi_{\mathrm{sem}}(p)+\gamma_2\psi_{\mathrm{str}}(p)
+\gamma_3\psi_{\mathrm{dist}}(p)
\label{eq:objective}
\end{equation}
where the weights \(\gamma_1\), \(\gamma_2\), and \(\gamma_3\) determine the relative contributions of the semantic, structural, and distance scores. The optimal placement is selected by restricting the search to feasible candidates:
\begin{equation}
\footnotesize
p^{*}=\underset{p\in\mathcal{P}_{\mathrm{feas}}}{\arg\max}\,\Phi(p)
\label{eq:best-position}
\end{equation}

\begin{figure*}[!tb]
\centering
\includegraphics[width=0.970\textwidth]{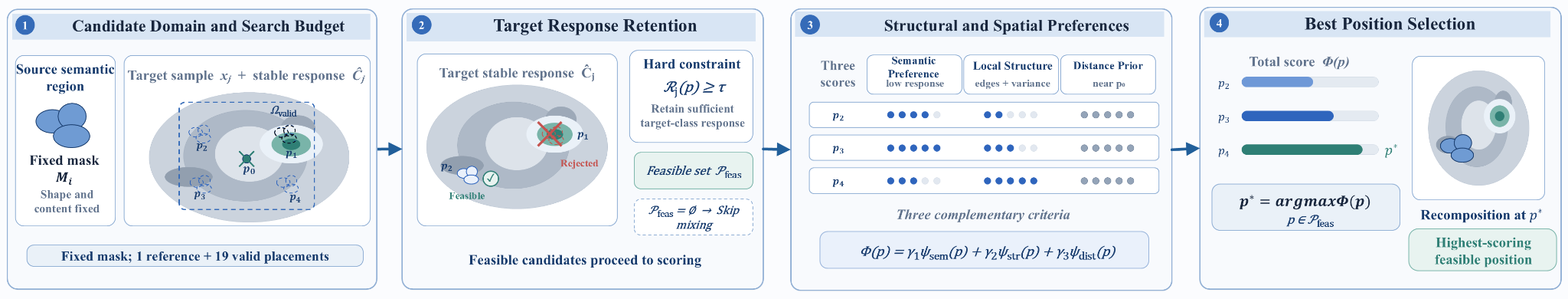}
\caption{Microscale searching for anchor-constrained two-dimensional placement under H2. The candidate budget consists of one reference placement at the original location and 19 placements uniformly sampled from the valid domain. The target-response retention constraint defines the feasible set. Semantic, structural, and distance scores are used to select \(p^{*}\) within this set. If the feasible set is empty, mixing is skipped.}
\label{fig:placement}
\end{figure*}

If \(\mathcal{P}_{\mathrm{feas}}\) is empty, mixing is skipped, and the original sample is retained. Figure~\ref{fig:placement} summarizes this constrained decision under H2.

\subsection{Mixed-Image Construction and Response-Based Supervision}
\label{sec:recombination}

Constrained placement alone does not fully define the mixed image and its supervision. The construction must address the intensity transition between source and target images, while the label must reflect the class evidence retained from each. Boundary smoothing and response-based label weights address these two aspects, thereby connecting the semantic anchor to the augmented sample.

\textbf{1) Mixed-Image Construction and Boundary Smoothing:} A hard binary boundary can introduce abrupt intensity transitions. Gaussian softening is applied to \(M_i^{p^{*}}\) to yield a continuous blending map \(A_i^{p^{*}}\), where \(\sigma\) controls the transition width. With the source content shifted to the optimal placement \(p^*\), the mixed image is constructed as

\begin{equation}
\footnotesize
\widetilde{x}=A_i^{p^{*}}\odot X_i^{p^{*}}+
\left(1-A_i^{p^{*}}\right)\odot x_j,
\label{eq:recombination}
\end{equation}
where \(\odot\) denotes element-wise multiplication. While boundary smoothing mitigates image-level transitions, the consistency loss \(\mathcal{L}_{\mathrm{cons}}\) penalizes feature-level discrepancies within the mixed region.

\textbf{2) Label Weights from Retained Responses:} A small transferred region can contain substantial class evidence. Therefore, its pixel area alone is not an adequate proxy for its contribution to supervision. The normalized stable-response maps are used to estimate the fraction of class evidence retained from each image. Let \(z\) index spatial locations in the corresponding image grid. The source fraction is

\begin{equation}
\footnotesize
r_i=\frac{\sum_z M_i(z)\widehat{C}_i(z)}
{\sum_z\widehat{C}_i(z)+\varepsilon}.
\label{eq:source-ratio}
\end{equation}

The retained target fraction is
\begin{equation}
\footnotesize
r_j=\frac{\sum_z[1-M_i^{p^{*}}(z)]\widehat{C}_j(z)}
{\sum_z\widehat{C}_j(z)+\varepsilon}.
\label{eq:target-ratio}
\end{equation}

The source mask \(M_i\) defines the transferred region, and its translated complement \(1-M_i^{p^{*}}\) defines the uncovered target region. These fractions are computed from the same regions used in the placement decision, ensuring consistency between placement and supervision. When the sum of the retained response fractions \(r_i+r_j\) exceeds a fixed threshold \(\delta>0\), the label weights are given by

\begin{equation}
\footnotesize
\lambda_i=\frac{r_i}{r_i+r_j},\qquad
\lambda_j=\frac{r_j}{r_i+r_j}.
\label{eq:label-weights}
\end{equation}
These weights sum to one. If the total response fraction is too small for normalization, area-based fractions are used as a fallback. The mixed label is

\begin{equation}
\footnotesize
\widetilde{y}=\lambda_i y_i+\lambda_j y_j,
\label{eq:mixed-label}
\end{equation}
where \(y_i\) and \(y_j\) are the one-hot labels defined in Section~\ref{sec:framework}. This response-based weighting ensures that the supervision signal reflects the class evidence retained from both images, aligning the label with the semantic constraints used during sample construction.

\section{Experiments}
\label{sec:experiments}

MSDA-LDSM is evaluated under unified data splits, backbones, and training budgets. Experiments are designed to assess the complete framework, test the two individual hypotheses, and compare MSDA-LDSM with existing augmentation methods.

\subsection{Experimental Settings}
\label{sec:settings}

We investigate the following four research questions:

\begin{samepage}
\begin{itemize}
 \item RQ1: How does MSDA-LDSM affect the accuracy and macro-F1 across the nine dataset-backbone settings compared with the baseline?
 \item RQ2: Can stochastic-response statistics identify repeatable local class evidence?
 \item RQ3: Can anchor-constrained placement produce effective mixed samples under a fixed candidate budget?
 \item RQ4: Does MSDA-LDSM outperform state-of-the-art augmentation methods?
\end{itemize}
\end{samepage}


\textbf{Datasets and splits:} We use ChestCT \cite{ref15}, BrainTumor \cite{ref16}, and BreakHis \cite{ref17}. The ChestCT collection contains 1,103 images from four classes and follows the class organization of the public Chest CT-Scan Images Dataset. BrainTumor contains 3,264 four-class brain MRI images, and BreakHis contains 2,013 breast histopathology images from eight subtypes at $200\times$ magnification. All methods use identical training, validation, and test splits. The validation set is used only for early stopping and model selection. The test set is reserved for final evaluation. The data splits are identical across all 10 runs, while the training seed varies per run.

\textbf{Implementation details:} Experiments use PyTorch with ResNet-18/50 \cite{ref18} and GoogLeNet \cite{szegedy2015googlenet}. The ResNet experiments use an NVIDIA GeForce RTX 2080 Ti. The common training protocol uses a batch size of 32 and permits at most 200 epochs. Training stops early if no validation improvement is observed for 50 epochs. Model and method settings are fixed across the 10 runs of each dataset-backbone combination.

\textbf{MSDA-LDSM parameters:} Dropout is placed in a high-level convolutional stage with probability $p=0.5$. We use $K=10$ stochastic responses, fusion coefficient $\eta=0.4$, foreground percentile $q=50$, and candidate budget $T_{\max}=20$. The high- and low-response thresholds are 0.8 and 0.3, respectively. The minimum semantic-region size is 64 pixels. The scoring weights are $\gamma_1=1.0$, $\gamma_2=0.3$, and $\gamma_3=0.2$. The minimum retention ratio is $\tau=0.8$. The Gaussian standard deviation is $\sigma=3.0$. The feature-consistency weight is $\lambda_1=0.1$.

\textbf{Comparison methods and metrics:} We compare the baseline, MixUp \cite{ref05}, CutMix \cite{ref06}, Co-Mixup \cite{ref22}, SaliencyMix \cite{ref21}, SnapMix \cite{ref07}, AlignMixup \cite{ref23}, MedAugment \cite{ref10}, SRA \cite{ref28}, HistAug-Adapter \cite{ref39}, StyCona-Cls \cite{ref35}, and MSDA-LDSM. Accuracy and macro-F1 are reported as mean \(\pm\) standard deviation over 10 runs. Wilcoxon signed-rank tests use identical seeds, with \(p<0.05\) indicating significance.

\textbf{Abbreviations:} CT, MRI, and BH denote ChestCT, BrainTumor, and BreakHis. R18 and R50 denote ResNet-18 and ResNet-50.

\subsection{RQ1: How Does MSDA-LDSM Affect Accuracy and Macro-F1 Across the Nine Dataset-Backbone Settings?}
\label{sec:rq1}

To answer RQ1, we compare the baseline with the full method under identical training settings. Table~\ref{tab:rq1} shows that MSDA-LDSM improves accuracy over the baseline in all nine dataset-backbone settings, with gains ranging from 0.86 to 16.12 percentage points. Macro-F1 improves in seven settings. In the two BrainTumor settings with ResNet-18/50, the differences from the baseline are $-0.30$ and $-1.07$ percentage points, respectively.

\emph{These results show consistent accuracy improvements across the evaluated modalities and backbones.}

\begin{table*}[!tb]
\centering
\scriptsize
\caption{Classification performance of Baseline and MSDA-LDSM on three medical image datasets (\%, mean $\pm$ std. dev. over 10 runs)}
\renewcommand{\arraystretch}{1.08}
\setlength{\tabcolsep}{2pt}
\begin{tabularx}{\textwidth}{@{}c l *{6}{Y}@{}}
\toprule
\textbf{Backbone} & \textbf{Method} & \textbf{CT Acc.} & \textbf{MRI Acc.} & \textbf{BH Acc.} & \textbf{CT Macro-F1} & \textbf{MRI Macro-F1} & \textbf{BH Macro-F1} \\
\midrule
\multirow{2}{*}{R18}
& Baseline
& \(74.51\pm3.41\)
& \(78.68\pm0.92\)
& \(84.69\pm2.99\)
& \(77.72\pm4.47\)
& \(\underline{75.38\pm1.90}\)
& \(82.55\pm2.75\) \\

& MSDA-LDSM
& \(\boldsymbol{90.63\pm2.27}\)
& \(\boldsymbol{82.95\pm1.27}\)
& \(\boldsymbol{92.83\pm0.62}\)
& \(\boldsymbol{91.10\pm2.14}\)
& \(75.08\pm2.14\)
& \(\boldsymbol{91.60\pm0.83}\) \\

\multirow{2}{*}{R50}
& Baseline
& \(73.81\pm2.58\)
& \(77.18\pm1.50\)
& \(86.54\pm0.57\)
& \(75.49\pm3.02\)
& \(\underline{73.77\pm0.21}\)
& \(83.18\pm1.96\) \\

& MSDA-LDSM
& \(\boldsymbol{88.43\pm0.94}\)
& \(\boldsymbol{81.68\pm0.45}\)
& \(\boldsymbol{91.59\pm1.80}\)
& \(\boldsymbol{84.53\pm3.04}\)
& \(72.70\pm1.67\)
& \(\boldsymbol{90.25\pm0.81}\) \\
\multirow{2}{*}{GoogLeNet} & Baseline & \(73.46\pm9.83\) & \(63.60\pm7.55\) & \(87.25\pm3.28\) & \(74.03\pm13.43\) & \(58.00\pm6.77\) & \(85.43\pm3.02\) \\
 & MSDA-LDSM & \(\boldsymbol{74.32\pm11.48}\) & \(\boldsymbol{79.29\pm1.28}\) & \(\boldsymbol{88.31\pm1.75}\) & \(\boldsymbol{74.38\pm13.85}\) & \(\boldsymbol{76.76\pm1.79}\) & \(\boldsymbol{85.48\pm2.56}\) \\
\midrule
\multicolumn{2}{l}{\textbf{\( +/=/- \)}} & \multicolumn{3}{c}{\(7/2/0\)} & \multicolumn{3}{c}{\(5/4/0\)} \\
\bottomrule
\end{tabularx}
\significancenote{* Counts summarize Wilcoxon signed-rank test at \(p<0.05\) across the nine dataset-backbone settings in the order \(+/=/-\). "\(+/=/-\)" denote significantly better performance, no significant difference, and significantly worse performance, respectively, relative to the comparison method. Best results are bold for MSDA-LDSM and underlined for other methods or variants.}
\label{tab:rq1}
\end{table*}

\subsection{RQ2: Can Stochastic-Response Statistics Identify Repeatable Local Class Evidence?}
\label{sec:rq2}

To answer RQ2, we evaluate the repeatability premise of H1 under fixed input images, target classes, and network parameters. The anchor-construction variants include a deterministic response (Single), a fusion of the deterministic response with the stochastic mean (Mean-fused), and the full formulation that additionally accounts for response variability (Stable, i.e., MSDA-LDSM).

For each dataset-backbone setting, we select the model with median validation performance from the 10 runs. For each of 10 test images per class, we collect 10 groups of \(K\) stochastic responses. To quantify repeatability, we evaluate the cross-group correlation of the continuous responses, the IoU of the resulting masks, the anchor-generation rate, and the downstream accuracy and macro-F1 (Tables~\ref{tab:h1a} and~\ref{tab:h1b}).

\begin{table*}[!tb]
\centering
\scriptsize
\caption{Response-repeatability and semantic-anchor stability metrics for H1 (mean \(\pm\) std. dev.)}
\renewcommand{\arraystretch}{1.08}
\setlength{\tabcolsep}{0pt}
\begin{tabularx}{\textwidth}{@{}l *{5}{Y}@{}}
\toprule
\textbf{Dataset-Backbone} & \textbf{\(\rho\) (individual stochastic response)} & \textbf{\(\rho(\text{Mean-fused})\)} & \textbf{\(\rho(\widehat{C})\)} & \textbf{\(\mathrm{IoU}(M)\)} & \textbf{Anchor-generation rate} \\
\midrule
CT-R18  & \(0.9912\pm0.0032\) & \(0.9997\pm0.0001\) & \(0.7771\pm0.1218\) & \(0.7058\pm0.0863\) & \(1.0000\pm0.0000\) \\
CT-R50  & \(0.9389\pm0.1474\) & \(0.9859\pm0.0781\) & \(0.9322\pm0.1328\) & \(0.7081\pm0.0515\) & \(0.9886\pm0.0357\) \\
MRI-R18 & \(0.9943\pm0.0037\) & \(0.9998\pm0.0001\) & \(0.7719\pm0.2005\) & \(0.7398\pm0.0697\) & \(1.0000\pm0.0000\) \\
MRI-R50 & \(0.7277\pm0.1396\) & \(0.9862\pm0.0099\) & \(0.8081\pm0.1266\) & \(0.7297\pm0.1488\) & \(0.8250\pm0.3389\) \\
BH-R18  & \(0.9931\pm0.0022\) & \(0.9997\pm0.0001\) & \(0.6493\pm0.1047\) & \(0.6233\pm0.0560\) & \(1.0000\pm0.0000\) \\
BH-R50  & \(0.8479\pm0.1864\) & \(0.8644\pm0.0565\) & \(0.7162\pm0.1037\) & \(0.6562\pm0.1981\) & \(0.7675\pm0.3346\) \\
\bottomrule
\end{tabularx}
\label{tab:h1a}
\end{table*}

\begin{table*}[!tb]
\centering
\scriptsize
\caption{Progressive controlled anchor experiment for H1 (\%, mean $\pm$ std. dev. over 10 runs)}
\renewcommand{\arraystretch}{1.08}
\setlength{\tabcolsep}{2pt}
\begin{tabularx}{\textwidth}{@{}*{8}{Y}@{}}
\toprule
\multicolumn{4}{c}{\textbf{Accuracy}} & \multicolumn{4}{c}{\textbf{Macro-F1}} \\
\midrule
\textbf{Setting} & \textbf{Single} & \textbf{Mean-fused} & \textbf{Stable} & \textbf{Setting} & \textbf{Single} & \textbf{Mean-fused} & \textbf{Stable} \\
CT-R18 & \(84.60\pm6.81\) & \(88.16\pm6.83\) & \(\boldsymbol{90.63\pm2.27}\) & CT-R18 & \(85.47\pm6.38\) & \(87.57\pm9.58\) & \(\boldsymbol{91.10\pm2.14}\) \\
CT-R50 & \(76.73\pm4.29\) & \(74.03\pm7.40\) & \(\boldsymbol{88.43\pm0.94}\) & CT-R50 & \(78.78\pm3.99\) & \(76.30\pm6.74\) & \(\boldsymbol{84.53\pm3.04}\) \\
MRI-R18 & \(77.74\pm0.57\) & \(77.97\pm0.66\) & \(\boldsymbol{82.95\pm1.27}\) & MRI-R18 & \(74.40\pm0.73\) & \(74.30\pm1.09\) & \(\boldsymbol{75.08\pm2.14}\) \\
MRI-R50 & \(73.86\pm4.66\) & \(73.25\pm3.01\) & \(\boldsymbol{81.68\pm0.45}\) & MRI-R50 & \(69.62\pm4.48\) & \(69.47\pm2.94\) & \(\boldsymbol{72.70\pm1.67}\) \\
BH-R18 & \(91.84\pm3.88\) & \(92.33\pm2.07\) & \(\boldsymbol{92.83\pm0.62}\) & BH-R18 & \(90.62\pm4.22\) & \(90.71\pm3.16\) & \(\boldsymbol{91.60\pm0.83}\) \\
BH-R50 & \(\underline{91.84\pm3.75}\) & \(88.88\pm5.80\) & \(91.59\pm1.80\) & BH-R50 & \(89.95\pm5.10\) & \(86.33\pm7.80\) & \(\boldsymbol{90.25\pm0.81}\) \\
CT-GNet & \(73.27\pm11.80\) & \(\underline{74.79\pm12.19}\) & \(74.32\pm11.48\) & CT-GNet & \(72.91\pm15.75\) & \(73.88\pm15.76\) & \(\boldsymbol{74.38\pm13.85}\) \\
MRI-GNet & \(60.76\pm7.87\) & \(62.11\pm7.08\) & \(\boldsymbol{79.29\pm1.28}\) & MRI-GNet & \(55.42\pm7.41\) & \(56.03\pm6.30\) & \(\boldsymbol{76.76\pm1.79}\) \\
BH-GNet & \(81.84\pm8.37\) & \(83.87\pm10.72\) & \(\boldsymbol{88.31\pm1.75}\) & BH-GNet & \(77.89\pm11.98\) & \(79.53\pm16.96\) & \(\boldsymbol{85.48\pm2.56}\) \\
\midrule
\textbf{\( +/=/- \)} & \(7/2/0\) & \(7/2/0\) & --- & \textbf{\( +/=/- \)} & \(7/2/0\) & \(7/2/0\) & --- \\
\bottomrule
\end{tabularx}
\significancenote{* Counts summarize Wilcoxon signed-rank test at \(p<0.05\) across the nine dataset-backbone settings in the order \(+/=/-\). "\(+/=/-\)" denote significantly better performance, no significant difference, and significantly worse performance, respectively, relative to the comparison method. Best results are bold for MSDA-LDSM and underlined for other methods or variants. GNet denotes GoogLeNet.}
\label{tab:h1b}
\end{table*}

\begin{figure}[!tbp]
\centering
\includegraphics[width=\columnwidth]{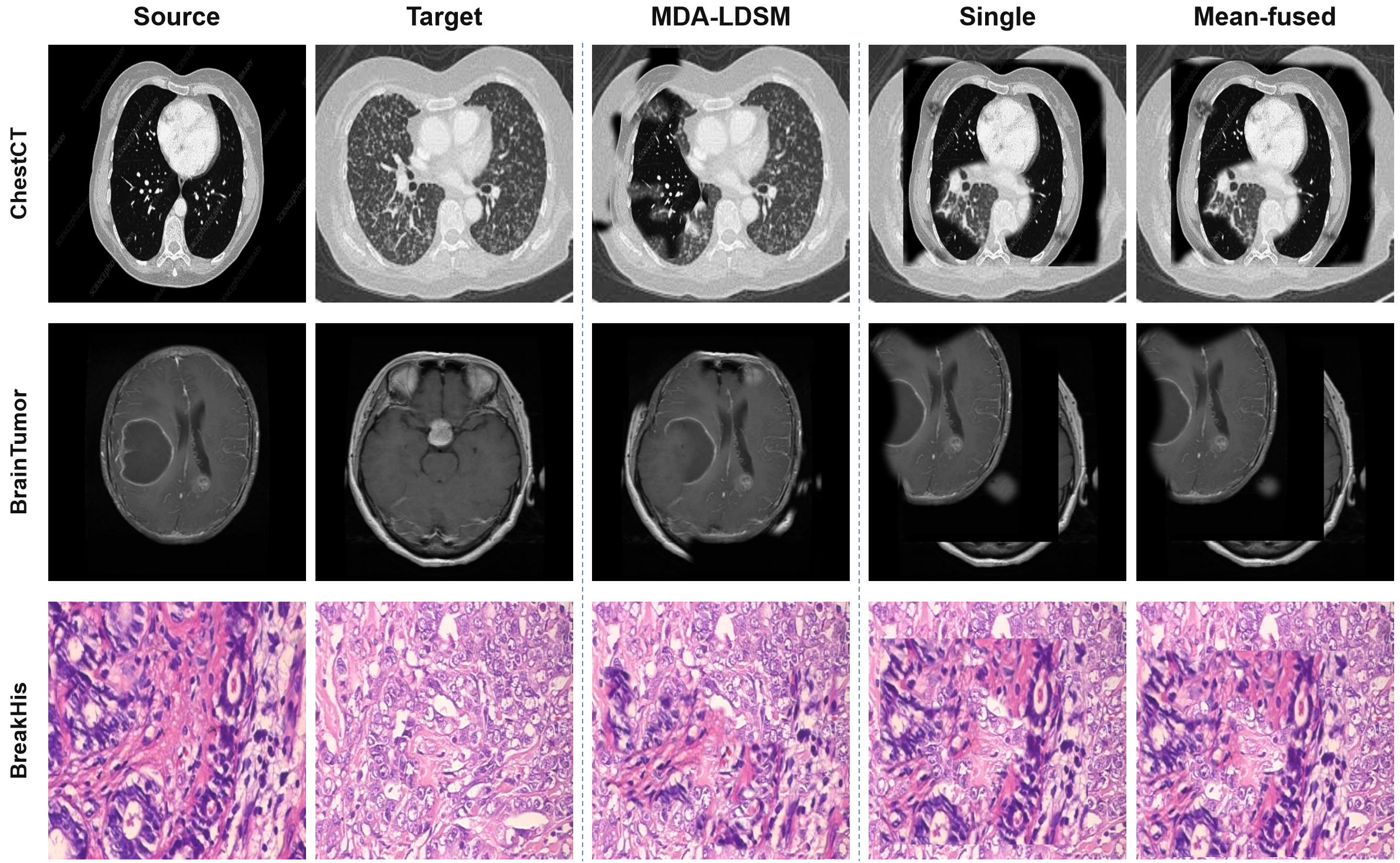}
\caption{Qualitative comparison of H1 anchor-construction variants. One source-target pair per dataset was selected to illustrate clear visual differences among variants. Columns show the source image, target image, and mixed samples from MSDA-LDSM, Single, and Mean-fused.}
\label{fig:h1-qualitative}
\end{figure}

As reported in Table~\ref{tab:h1a}, the response maps produced by the Stable variant have an average cross-group correlation of 0.7758, and their resulting masks have an average IoU of 0.6938. The Mean-fused variant achieves a higher average correlation of 0.9726. This suggests that the added variability weighting in the Stable variant shifts the spatial emphasis rather than maximizing correlation alone. Anchor-generation rates range from 76.75\% to 100\%. Figure~\ref{fig:h1-qualitative} qualitatively compares the mixed samples produced by the three anchor-construction variants. It shows that MSDA-LDSM yields smoother boundaries and better semantic coherence than the ablated alternatives.

On average across the nine settings, the Stable variant improves accuracy by 6.39 percentage points over the Single variant and by 6.07 percentage points over the Mean-fused variant. It also increases macro-F1 by 5.20 percentage points over the Single variant and by 5.31 percentage points over the Mean-fused variant.

\begin{samepage}
\emph{These findings support H1: response intensity and variability are jointly used to identify repeatable local class evidence for semantic-anchor construction.}
\end{samepage}

\subsection{RQ3: Can Anchor-Constrained Placement Produce Effective Mixed Samples Under a Fixed Candidate Budget?}
\label{sec:rq3}

To answer RQ3, we compare placement variants using retention rejection rates, sample-generation rates, and downstream classification performance. The variants include a reference placement at the original location, a no-retention variant, and a random-feasible placement variant. The reference variant only evaluates \(p_0\) subject to the retention filter. The no-retention variant retains the full candidate set and the scoring objective but removes the retention constraint. The random-feasible placement variant retains the retention filter but selects a feasible candidate uniformly at random. The full method considers one reference placement and 19 uniformly sampled valid placements under the budget \(T_{\max}=20\). All variants share the same non-placement components and training settings.

For the placement-process evaluation, we select the model with median validation performance and use the same 1,000 source-target pairs in each setting. For each variant, we record the retention rejection rate and the sample-generation rate. We also record the semantic, structural, and distance scores of the selected placements. Downstream comparisons use identical data splits and 10 paired seeds.

\begin{table*}[!tb]
\centering
\scriptsize
\caption{Generation process and constraint satisfaction of the core H2 variants (mean \(\pm\) std. dev.)}
\renewcommand{\arraystretch}{1.08}
\setlength{\tabcolsep}{2pt}
\begin{tabularx}{\textwidth}{@{}L *{5}{Y}@{}}
\toprule
\textbf{Variant} & \textbf{CAM rejection rate} & \textbf{Sample-generation rate} & \textbf{\(\mu_{\mathrm{CAM}}(p^{*})\)} & \textbf{\(\sigma_{\mathrm{gray}}(p^{*})\)} & \textbf{\(d(p^{*},p_0)\)} \\
\midrule
\textbf{MSDA-LDSM}        & \(0.0801\pm0.2633\) & \(0.9352\pm0.2462\) & \(0.0947\pm0.0922\) & \(0.1305\pm0.0552\) & \(0.0198\pm0.0524\) \\
\textbf{Same-position Only} & \(0.0837\pm0.2769\) & \(0.9163\pm0.2769\) & \(0.0962\pm0.0928\) & \(0.1303\pm0.0552\) & \(0.0022\pm0.0006\) \\
\textbf{w/o Hard Filtering} & --- & \(1.0000\pm0.0000\) & \(0.1114\pm0.1097\) & \(0.1311\pm0.0550\) & \(0.0203\pm0.0513\) \\
\textbf{Random Feasible Position} & \(0.0801\pm0.2633\) & \(0.9352\pm0.2462\) & \(0.0983\pm0.0949\) & \(0.1309\pm0.0553\) & \(0.0763\pm0.0958\) \\
\bottomrule
\end{tabularx}
\par\vspace{2pt}\noindent\parbox{\textwidth}{\scriptsize ``---'' indicates that the CAM rejection rate is not applicable because this variant disables CAM hard filtering.}
\label{tab:h2a}
\end{table*}

\begin{table*}[!tb]
\centering
\scriptsize
\caption{Controlled classification experiment for the core H2 placement-search variants (mean \(\pm\) std. dev. over 10 runs)}
\renewcommand{\arraystretch}{1.08}
\setlength{\tabcolsep}{0pt}
\begin{tabularx}{\textwidth}{@{}>{\centering\arraybackslash}p{34pt}*{4}{Y}>{\centering\arraybackslash}p{34pt}*{4}{Y}@{}}
\toprule
\multicolumn{5}{c}{\textbf{Accuracy}} & \multicolumn{5}{c}{\textbf{Macro-F1}} \\
\midrule
\textbf{Setting} & \textbf{Full} & \textbf{Same-pos.} & \textbf{w/o filter} & \textbf{Random} & \textbf{Setting} & \textbf{Full} & \textbf{Same-pos.} & \textbf{w/o filter} & \textbf{Random} \\
CT-R18 & \(\boldsymbol{90.63\pm2.27}\) & \(80.79\pm6.00\) & \(90.03\pm3.45\) & \(68.67\pm6.78\) & CT-R18 & \(\boldsymbol{91.10\pm2.14}\) & \(82.36\pm5.90\) & \(90.44\pm3.27\) & \(69.34\pm8.85\) \\
CT-R50 & \(\boldsymbol{88.43\pm0.94}\) & \(75.08\pm3.55\) & \(76.38\pm6.32\) & \(77.05\pm5.87\) & CT-R50 & \(\boldsymbol{84.53\pm3.04}\) & \(77.58\pm3.35\) & \(78.62\pm5.96\) & \(78.41\pm6.60\) \\
MRI-R18 & \(\boldsymbol{82.95\pm1.27}\) & \(79.04\pm1.03\) & \(77.66\pm1.11\) & \(73.32\pm7.68\) & MRI-R18 & \(75.08\pm2.14\) & \(\underline{76.01\pm1.42}\) & \(73.93\pm1.74\) & \(70.23\pm7.41\) \\
MRI-R50 & \(\boldsymbol{81.68\pm0.45}\) & \(75.76\pm2.30\) & \(76.19\pm1.91\) & \(76.04\pm1.80\) & MRI-R50 & \(72.70\pm1.67\) & \(72.00\pm3.13\) & \(\underline{72.91\pm2.56}\) & \(72.44\pm1.65\) \\
BH-R18 & \(\boldsymbol{92.83\pm0.62}\) & \(90.77\pm5.01\) & \(90.47\pm6.25\) & \(66.33\pm9.60\) & BH-R18 & \(\boldsymbol{91.60\pm0.83}\) & \(89.32\pm5.60\) & \(88.36\pm8.49\) & \(51.47\pm18.39\) \\
BH-R50 & \(\boldsymbol{91.59\pm1.80}\) & \(80.89\pm17.12\) & \(90.05\pm3.83\) & \(86.43\pm7.62\) & BH-R50 & \(\boldsymbol{90.25\pm0.81}\) & \(72.48\pm28.90\) & \(88.01\pm4.68\) & \(83.54\pm9.36\) \\
CT-GNet & \(74.32\pm11.48\) & \(\underline{77.05\pm9.95}\) & \(74.48\pm8.31\) & \(71.40\pm10.60\) & CT-GNet & \(74.38\pm13.85\) & \(\underline{77.87\pm12.00}\) & \(76.14\pm9.31\) & \(70.87\pm14.41\) \\
MRI-GNet & \(\boldsymbol{79.29\pm1.28}\) & \(63.68\pm6.21\) & \(68.20\pm1.89\) & \(58.02\pm4.94\) & MRI-GNet & \(\boldsymbol{76.76\pm1.79}\) & \(59.29\pm4.65\) & \(61.74\pm2.59\) & \(52.48\pm5.11\) \\
BH-GNet & \(88.31\pm1.75\) & \(87.05\pm2.11\) & \(\underline{89.88\pm1.34}\) & \(83.15\pm5.24\) & BH-GNet & \(85.48\pm2.56\) & \(84.61\pm3.08\) & \(\underline{88.56\pm1.39}\) & \(81.00\pm6.05\) \\
\midrule
\textbf{\( +/=/- \)} & --- & \(6/3/0\) & \(6/2/1\) & \(8/1/0\) & \textbf{\( +/=/- \)} & --- & \(6/3/0\) & \(5/3/1\) & \(7/2/0\) \\
\bottomrule
\end{tabularx}
\significancenote{* Counts summarize Wilcoxon signed-rank test at \(p<0.05\) across the nine dataset-backbone settings in the order \(+/=/-\). "\(+/=/-\)" denote significantly better performance, no significant difference, and significantly worse performance, respectively, relative to the comparison method. Best results are bold for MSDA-LDSM and underlined for other methods or variants. GNet denotes GoogLeNet.}
\label{tab:h2b}
\end{table*}

As reported in Table~\ref{tab:h2a}, the retention filter rejects 8.01\% of the sampled candidates in the full method. The sample-generation rate is 93.52\%, compared with 91.63\% for the reference variant. The mean target response at the selected placement is 0.0947 for the full method and 0.0983 for the random-feasible variant. The lower mean target response aligns with the semantic score in \eqref{eq:semantic-score}. The corresponding mean normalized distances are 0.0198 and 0.0763, respectively, confirming the intended distance preference toward the reference coordinate.

Compared with the reference variant, the no-retention variant, and the random-feasible variant, the full method improves average accuracy by 6.66, 4.08, and 12.18 percentage points, respectively. The corresponding macro-F1 gains are 5.60, 2.57, and 12.46 percentage points.

\emph{These findings support H2 by showing that a fixed semantic region with constrained two-dimensional placement enables effective mixed-sample augmentation under a limited candidate budget.}

\subsection{RQ4: Does MSDA-LDSM Outperform State-of-the-Art Augmentation Methods?}
\label{sec:rq4}

For ResNet-18/50, Table~\ref{tab:comparison} presents a comparison of MSDA-LDSM with general mixed-sample, medical-image-specific, and sample-adaptive augmentation methods under unified splits and training budgets. MSDA-LDSM attains the best mean rank for accuracy (1.33). Its macro-F1 rank varies across dataset-backbone settings.

\begin{table*}[!tb]
\centering
\scriptsize
\caption{Accuracy and Macro-F1 comparison of MSDA-LDSM and existing augmentation methods on three medical image datasets (mean \(\pm\) std. dev. over 10 runs)}
\renewcommand{\arraystretch}{1.05}
\setlength{\tabcolsep}{1.7pt}

\vspace{-0.5em}
\textbf{(a) Accuracy}

\begin{tabularx}{\textwidth}{@{}L *{6}{Y} c c@{}}
\toprule
\textbf{Method} & \textbf{CT-R18} & \textbf{CT-R50} & \textbf{MRI-R18} & \textbf{MRI-R50} & \textbf{BH-R18} & \textbf{BH-R50} & \textbf{Avg. Rank} & \textbf{\(+/=/-\)} \\
\midrule
Baseline        & \(74.51\pm3.41\) & \(73.81\pm2.58\) & \(78.68\pm0.92\) & \(77.18\pm1.50\) & \(84.69\pm2.99\) & \(86.54\pm0.57\) & 6.67 & \(6/0/0\) \\
MixUp           & \(58.28\pm4.02\) & \(57.30\pm2.56\) & \(78.87\pm1.01\) & \(76.65\pm1.39\) & \(75.04\pm0.24\) & \(78.91\pm1.73\) & 9.67 & \(6/0/0\) \\
CutMix          & \(73.84\pm2.76\) & \(71.19\pm1.80\) & \(77.35\pm1.11\) & \(76.90\pm1.42\) & \(83.01\pm2.13\) & \(85.30\pm2.62\) & 8.50 & \(6/0/0\) \\
Co-Mixup        & \(42.34\pm6.32\) & \(77.62\pm4.76\) & \(77.98\pm0.99\) & \(76.52\pm1.10\) & \(54.28\pm9.30\) & \(83.93\pm0.83\) & 9.50 & \(6/0/0\) \\
SaliencyMix     & \(69.56\pm6.94\) & \(60.87\pm7.33\) & \(76.78\pm0.28\) & \(74.56\pm0.52\) & \(72.93\pm4.68\) & \(70.72\pm1.73\) & 10.67 & \(6/0/0\) \\
SnapMix         & \(79.59\pm2.72\) & \(74.76\pm0.91\) & \(78.81\pm1.22\) & \(78.05\pm0.91\) & \(86.01\pm3.51\) & \(87.22\pm2.28\) & 5.00 & \(6/0/0\) \\
AlignMixup      & \(77.46\pm1.66\) & \(71.90\pm0.53\) & \(78.93\pm1.15\) & \(76.27\pm0.63\) & \(87.22\pm1.84\) & \(85.48\pm1.19\) & 6.50 & \(6/0/0\) \\
MedAugment      & \(86.98\pm3.62\) & \(79.84\pm5.64\) & \(80.94\pm1.11\) & \(79.95\pm1.52\) & \(88.09\pm2.65\) & \(86.77\pm4.15\) & 3.50 & \(6/0/0\) \\
SRA             & \(\underline{90.86\pm1.87}\) & \(84.54\pm3.59\) & \(82.23\pm1.01\) & \(\underline{81.80\pm0.76}\) & \(90.15\pm0.55\) & \(90.72\pm0.96\) & \(1.67\) & \(4/2/0\) \\
HistAug-Adapter & \(66.83\pm17.30\) & \(71.83\pm1.50\) & \(79.31\pm0.25\) & \(77.41\pm0.72\) & \(85.42\pm1.53\) & \(85.36\pm1.15\) & 7.00 & \(6/0/0\) \\
StyCona-Cls     & \(70.92\pm6.19\) & \(59.71\pm4.29\) & \(81.17\pm0.64\) & \(80.76\pm0.61\) & \(67.77\pm2.28\) & \(68.21\pm1.18\) & 8.00 & \(6/0/0\) \\
MSDA-LDSM       & \(90.63\pm2.27\) & \(\boldsymbol{88.43\pm0.94}\) & \(\boldsymbol{82.95\pm1.27}\) & \(81.68\pm0.45\) & \(\boldsymbol{92.83\pm0.62}\) & \(\boldsymbol{91.59\pm1.80}\) & \(\boldsymbol{1.33}\) & --- \\
\bottomrule
\end{tabularx}

\vspace{0.5em}

\textbf{(b) Macro-F1}

\begin{tabularx}{\textwidth}{@{}L *{6}{Y} c c@{}}
\toprule
\textbf{Method} & \textbf{CT-R18} & \textbf{CT-R50} & \textbf{MRI-R18} & \textbf{MRI-R50} & \textbf{BH-R18} & \textbf{BH-R50} & \textbf{Avg. Rank} & \textbf{\(+/=/-\)} \\
\midrule
Baseline        & \(77.72\pm4.47\) & \(75.49\pm3.02\) & \(75.38\pm1.90\) & \(73.77\pm0.21\) & \(82.55\pm2.75\) & \(83.18\pm1.96\) & 6.67 & \(4/2/0\) \\
MixUp           & \(67.70\pm2.85\) & \(45.85\pm2.60\) & \(75.22\pm1.81\) & \(74.57\pm0.12\) & \(75.30\pm0.71\) & \(75.41\pm0.19\) & 9.33 & \(4/1/1\) \\
CutMix          & \(77.27\pm1.58\) & \(74.94\pm0.35\) & \(73.44\pm3.18\) & \(74.07\pm1.39\) & \(87.39\pm0.80\) & \(83.48\pm1.87\) & 6.33 & \(5/0/1\) \\
Co-Mixup        & \(78.85\pm2.15\) & \(80.75\pm5.10\) & \(76.33\pm0.59\) & \(70.44\pm0.52\) & \(83.91\pm0.70\) & \(81.81\pm3.90\) & 6.67 & \(5/1/0\) \\
SaliencyMix     & \(69.44\pm8.83\) & \(71.66\pm1.67\) & \(70.18\pm0.61\) & \(65.30\pm3.92\) & \(65.62\pm6.58\) & \(65.17\pm5.20\) & 10.83 & \(6/0/0\) \\
SnapMix         & \(81.00\pm9.45\) & \(81.80\pm0.64\) & \(77.72\pm3.11\) & \(72.41\pm0.04\) & \(86.84\pm1.61\) & \(79.11\pm11.74\) & 5.50 & \(4/1/1\) \\
AlignMixup      & \(80.20\pm1.51\) & \(70.02\pm2.93\) & \(75.63\pm0.66\) & \(72.97\pm1.93\) & \(85.93\pm6.42\) & \(77.23\pm4.91\) & 7.17 & \(4/2/0\) \\
MedAugment      & \(87.82\pm3.44\) & \(81.19\pm5.31\) & \(78.97\pm1.43\) & \(77.99\pm1.43\) & \(84.47\pm3.64\) & \(83.77\pm5.82\) & 3.67 & \(4/0/2\) \\
SRA             & \(\underline{91.41\pm1.85}\) & \(\underline{85.54\pm3.15}\) & \(\underline{80.43\pm1.41}\) & \(\underline{79.38\pm0.97}\) & \(87.36\pm0.98\) & \(88.46\pm1.26\) & \(\underline{1.50}\) & \(2/2/2\) \\
HistAug-Adapter & \(63.28\pm27.35\) & \(75.07\pm1.21\) & \(76.28\pm1.35\) & \(72.74\pm0.94\) & \(81.10\pm2.04\) & \(81.62\pm1.51\) & 8.17 & \(4/2/0\) \\
StyCona-Cls     & \(72.21\pm5.78\) & \(58.84\pm6.18\) & \(79.19\pm0.84\) & \(78.83\pm0.53\) & \(57.37\pm2.09\) & \(57.49\pm2.10\) & 8.00 & \(4/0/2\) \\
MSDA-LDSM       & \(91.10\pm2.14\) & \(84.53\pm3.04\) & \(75.08\pm2.14\) & \(72.70\pm1.67\) & \(\boldsymbol{91.60\pm0.83}\) & \(\boldsymbol{90.25\pm0.81}\) & \(4.17\) & --- \\
\bottomrule
\end{tabularx}

\significancenote{* Counts summarize Wilcoxon signed-rank test at \(p<0.05\) across the six dataset-backbone settings in the order \(+/=/-\). "\(+/=/-\)" denote significantly better performance, no significant difference, and significantly worse performance, respectively, relative to the comparison method. Best results are bold for MSDA-LDSM and underlined for other methods or variants. Lower average ranks are better.}
\label{tab:comparison}
\end{table*}

Across the six settings, MSDA-LDSM achieves an average accuracy rank of 1.33, with the highest mean accuracy in CT-R50, MRI-R18, BH-R18, and BH-R50. Its gains over the baseline range from 4.27 to 16.12 percentage points, showing improvements on every dataset-backbone combination. For macro-F1, SRA attains the best average rank of 1.50 and leads in all ChestCT and BrainTumor settings. MSDA-LDSM achieves the best results on BH-R18 and BH-R50, exceeding SRA by 4.24 and 1.79 percentage points, respectively. It ranks second in both ChestCT settings. The best result in each column is bold for MSDA-LDSM and underlined for a comparison method. Compared with SRA, MSDA-LDSM achieves significantly higher accuracy in CT-R50, MRI-R18, BH-R18, and BH-R50. SRA has slightly higher mean accuracy in CT-R18 and MRI-R50, by 0.23 and 0.12 percentage points, respectively, but neither difference is statistically significant.

Across the six dataset-backbone settings, Friedman tests indicate overall differences among the 12 methods for accuracy $(\chi^2(11)=48.10,\ p<0.001)$ and macro-F1 $(\chi^2(11)=33.15,\ p<0.001)$.

\emph{MSDA-LDSM achieves the best mean accuracy rank. It consistently improves accuracy over the baseline across all six dataset-backbone settings. For macro-F1, it attains the best results on both BreakHis settings and remains competitive with state-of-the-art methods overall.}

\section{Conclusion}
\label{sec:conclusion}

This paper presents MSDA-LDSM for tumor classification in medical images using image-level class labels. The framework integrates variability-aware source-region selection, constrained two-dimensional placement, and response-based supervision. Ablation studies confirm that incorporating response variability for anchor construction (H1) and applying constrained placement (H2) consistently yield higher average accuracy than their respective variants. Mixed-label weights are derived from the response fractions retained after region transfer. Experiments on ChestCT, BrainTumor, and BreakHis with ResNet-18/50 and GoogLeNet demonstrate consistent accuracy gains over the baseline across all nine settings. MSDA-LDSM also attains the best mean accuracy rank in the external comparison. These results validate the combined use of variability-aware anchors and constrained placement for local mixing without pixel-level lesion annotations.

Future work will examine generalization across additional centers, imaging conditions, and model architectures.

\bibliographystyle{IEEEtran}
\bibliography{references}

\end{document}